# Nepali Video Captioning using CNN-RNN Architecture

Bipesh Subedi
Information and Language Processing
Research Lab (ILPRL) Lab
Department of Computer Science and Engg.
Kathmandu University, Dhulikhel,Nepal
bipeshrajsubedi@gmail.com

Saugat Singh
Information and Language Processing
Research Lab (ILPRL) Lab
Department of Computer Science and Engg.
Kathmandu University, Dhulikhel,Nepal
saugat.singh09@gmail.com

Bal Krishna Bal
Information and Language Processing
Research Lab (ILPRL) Lab
Department of Computer Science and Engg.
Kathmandu University, Dhulikhel,Nepal
bal@ku.edu.np

*Abstract*—This article presents a study on Nepali video captioning using deep neural networks. Through the integration of pre-trained CNNs and RNNs, the research focuses on generating precise and contextually relevant captions for Nepali videos. The approach involves dataset collection, data preprocessing, model implementation, and evaluation. By enriching the MSVD dataset with Nepali captions via Google Translate, the study trains various CNN-RNN architectures. The research explores the effectiveness of CNNs (e.g., EfficientNetB0, ResNet101, VGG16) paired with different RNN decoders like LSTM, GRU, and BiLSTM. Evaluation involves BLEU and METEOR metrics, with the best model being EfficientNetB0 + BiLSTM with 1024 hidden dimensions, achieving a BLEU-4 score of 17 and METEOR score of 46. The article also outlines challenges and future directions for advancing Nepali video captioning, offering a crucial resource for further research in this area.

## I. INTRODUCTION

Video captioning involves automatically creating concise and accurate textual descriptions (captions) for a video's visual content, enabling viewers to understand the context without audio. Video captioning plays a crucial role in fostering inclusivity for individuals with hearing impairments, enabling them to fully interact with video content. It also facilitates language translation, improving accessibility for a global audience. Captions enhance learning experiences by aiding comprehension and retention in educational videos, boost content discoverability and search engine rankings, and add visual reinforcement to spoken content, creating an engaging viewing experience. Overall, video captioning is a vital tool that promotes inclusivity, bridges communication gaps, and enhances multimedia experiences. Most popular method involves extracting visual features from video frames using CNNs and generating captions based on this information through RNN-based language models. The model is trained on a dataset of video-caption pairs to produce coherent and contextually appropriate captions for new videos. Other approaches for video captioning such as sequence-to-sequence, transformers, and audio-visual also exist.

When comparing video captioning to image captioning, both aim to provide textual annotations for visual content. However, video captioning considers temporal dynamics and handles dynamic content from video frames, while image captioning focuses on static images. Despite these differences, both use similar deep learning models, such as CNNs and RNNs, to improve visual content's usability, accessibility, and comprehension for diverse audiences. Within video captioning, two methods exist: dense captioning, which provides multiple captions for different segments, offering thorough context, and single captioning, which offers a concise summary of the entire video. In this work, CNN-RNN based encoder-decoder model is implemented for Nepali video captioning for single captions.

This study addresses the dearth of research and resources dedicated to producing accurate and contextually appropriate single captions for Nepali videos. Our approach involves employing a CNN-RNN based encoder-decoder model to generate succinct yet comprehensive summaries for such videos. Through this, we aim to enhance accessibility, inclusivity, and educational experiences. Additionally, we provide a curated Nepali dataset[1] to facilitate advancements in this domain.

## II. RELATED WORKS

Video captioning is a crucial area of research that focuses on automatically generating textual descriptions for the visual content of videos, thereby enhancing comprehension and accessibility. Over the years, numerous studies have been conducted to explore different approaches and models to improve the quality of video captions in regards with the grammatical correctness, sequence matching, and performance scores. Video captioning research has seen significant advancements over the years. With the introduction of LSTM model called S2VT [1], that experimented with different datasets like M-VAD, MPII_MD, and MSVD with feature extraction and flow extraction using VGG and AlexNet, achieved an impressive METEOR score of 29.8 for the MSVD dataset. This model has since become a benchmark for other researchers in video captioning. Another attempt to improve the study, adding the Flickr dataset for training on the MSVD dataset , for the CNN and LSTM models, resulted in increased BLEU-4 and METEOR scores [2]. Clip4Captions [3], an ensemble models with CLIP as encoder and BERT as decoder model outperformed regular models on testing .

Regarding the choice of dataset, a study used the MSR-VTT dataset for their CNN+LSTM based model, achieving a BLEU-4 score of 41 and METEOR of 28.7 [4]. Contradictorily, another study [5] found better results on the MSVD dataset than on MSR-VTT for their experiments, experimenting with LSTM, Bi-LSTM, and Reinforced Bi-LSTM models similar to SV2T study [1] . And they managed to obtain better results with the ReBiLSTM model, achieving a BLEU-4 score of 37.3 and METEOR of 30.3, outperforming the benchmarked S2VT models. On the other hand, training their CNN+RNN based model on the COCO dataset,the study obtained a BLEU score of 85.2 [6]. This clearly indicates that the choice of dataset doesn't truly affect the performance.

---

[1]https://www.kaggle.com/datasets/bipeshrajsubedi/msvd-nepali-dataset

Different pretrained models have also been explored. a study used the MSVD dataset for their CNN+LSTM based architecture and found Inception-v3 to be the preferred pretrained model over VGG16, ResNet152, and GoogLeNet [7]. Similarly, a study done on 2023 [8] reported that NASNet-Large yielded better scores with a BLEU-4 score of 50.3 for their CNN-LSTM based model. However on the same year, VGG16 used for feature extraction gave the best results for their experiments [9]. The different pretrained CNN models experimented by the studies indicates that there is no one-size-fits-all approach, and the effectiveness of pretrained models depends on the specific task and dataset characteristics.

The recent trends to investigate attention mechanisms to improve performance have shown promising results. Study on attention mechanism for their CNN+LSTM model [10], achieved a METEOR score of 32.1 and BLEU-4 score of 47.3, leading to performance improvement. Furthermore, multi-layer attention proposed by Naik [11] also showed improved performance on their RNN(LSTM,BiLSTM) based model.

While video captioning research has progressed, the published works are still limited.A study done on simililar low resource language explored the use of translated MSR-VTT dataset in Hindi for their 3DCNN+LSTM based model [12], achieving a METEOR score of 39.3. A study on image captioning in Nepali, found EfficientNet to be good option for the feature extraction [13].

By synthesizing and comparing the findings from different studies, it becomes evident that video captioning research continues to evolve, with various models, datasets, and attention mechanisms contributing to advancements in the field. Future research directions should focus on addressing the dataset bias, exploring novel architectures, and incorporating diverse languages to make video captioning accessible to a broader audience.

Our approach builds upon these works by combining feature extraction from pre-trained CNN models(EfficientNetB0, ResNet101, VGG16) with an Encoder-Decoder RNN (LSTM, BiLSTM, GRU) models. This allows us to leverage the strengths of both CNNs and RNNs for video captioning.

### III. METHODOLOGY

The methodology for this work involves a series of procedures, as illustrated in Fig. 1.

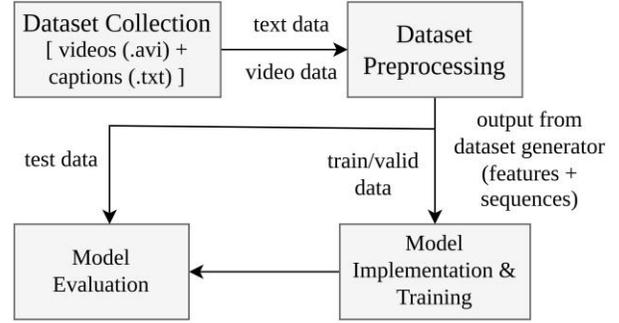

Fig. 1. High level flow of the procedures

#### A. Dataset Collection

The study utilized the MSVD[2] dataset published by University of Texas at Austin and Microsoft Research Natural Language Processing Group [14], comprising 1970 distinct videos having more than 80k text descriptions with an average of around 40 captions per video. It is important to note that the number of captions in each video varies. The text description contains video id and english captions which are processed further to make it compatible for this work.

#### B. Dataset Preprocessing

The raw dataset cannot be directly used for this work, as we are dealing with the Nepali language, and the neural network requires a specific shape of input data, therefore it is important to perform some preprocessing steps for both text and video data.

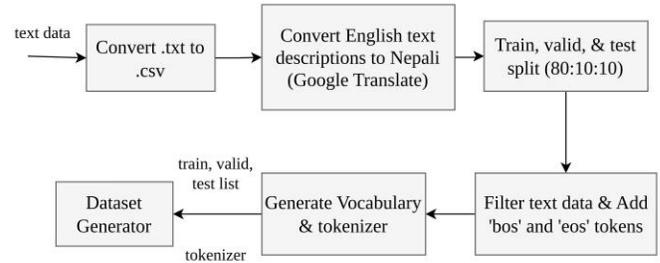

Fig. 2. Text preprocessing procedures

*1) Text Preprocessing:* Initially, the text file is transformed into a CSV file with columns for video ID, original text, and English captions. The video ID and English captions are then extracted from the original text. The next step involves translating the English text to Nepali, which is accomplished using the Google Translate API[3]. Subsequently, the translated Nepali captions are added to the CSV file. To prepare the data for training, the text is split into train, validation, and test data frames in an 80:10:10 ratio. Each caption is enriched by adding a "beginning of sentence" ('bos') token at the start and an "end of sentence" ('eos') token at the end. Additionally, captions with word frequency less than 2 are filtered out to ensure higher quality data. The resulting captions and video IDs are organized into separate training, validation, and test lists. The vocabulary is constructed exclusively from the captions in the training list since we will tokenize using only the words from the training data. The vocabulary comprises 15,585 unique words and is tokenized using the Keras Tokenizer[4]. After tokenization, the captions are padded to ensure that all sentences have the same length. To determine the maximum caption length, the average caption length of the dataset is calculated, which is found to be 5. However, some captions exceed a length of 35 words. Considering the potential challenges of dealing with excessive padding and the risk of the model overpredicting the padded tokens, a maximum caption length of 10 is chosen. The actual tokenization and padding of captions occurs during the dataset generation when creating the training and validation sequences. Fig. 2 depicts all the procedures involved in this step.

[2]https://www.cs.utexas.edu/users/ml/clamp/videoDescription/
[3]https://py-googletrans.readthedocs.io/en/latest/

features. Therefore for 30 frames, the features would be an array of shape (30, 1280), (30, 2094), or (30, 4096) depending upon the CNN model used. To extract those features, the final classification layer of the CNN model is removed. The overall process followed for this step is shown in Fig. 3.

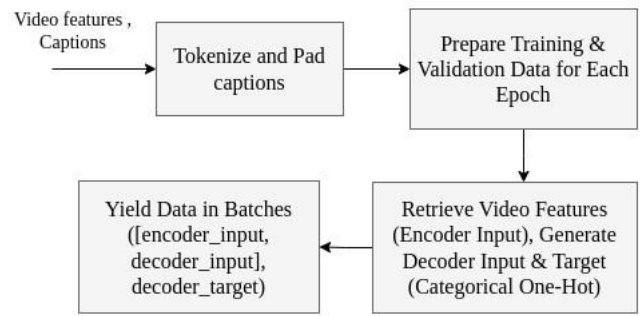

Fig. 4. Procedures involved in dataset generator

3) Dataset Generator: After preprocessing the videos and captions, they are fed into the model for training and validation using a dataset generator. Fig. 4 depicts the major steps involved in the dataset generator. This generator plays a crucial role in preparing and organizing the data for the video captioning model. It first separates the video IDs from their corresponding captions and converts the captions into tokens. Then, it pads the tokenized captions to ensure equal lengths for consistent processing. The generator creates training and validation data for each epoch and each caption in training and validation sequences. For each caption, it retrieves the video features (encoder input data) using the corresponding video ID. Additionally, it generates decoder input and target data by converting the caption tokens into categorical one-hot encoded vectors. The input sequence excludes the last token, while the target sequence excludes the first token. The generator efficiently manages the data in batches, optimizing memory usage. This approach enables the model to effectively learn from video-caption pairs while efficiently handling large datasets.

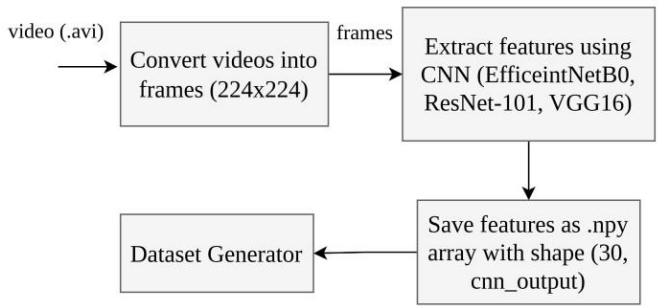

Fig. 3. Video preprocessing procedures

2) Video Preprocessing: Frames are individual images that make up a video, and they are extracted for various purposes. By analyzing frames, we can extract visual features and patterns using computer vision techniques like CNNs. Additionally, analyzing frames in sequence allows us to capture temporal information, such as movement and changes over time. In this work, frames (224x224 pixels) are extracted using OpenCV library and then features are extracted from these frames mainly using pre-trained CNN models: EfficientNetB0, ResNet101, and VGG16. In video captioning, a common method is to select a fixed number of frames evenly from the video to represent its content and generate captions. In our study, we chose 30 frames for each video because it creates a balance between capturing enough visual information to describe the content adequately while avoiding excessive computational overhead. The number of features extracted varies based on the CNN model used: EfficientNetB0 has 1280 features, ResNet101 has 2048 features, and VGG16 has 4096

C. Model Implementation and Training

The model architecture for video captioning explores various combinations of Convolutional Neural Networks (CNNs) and Recurrent Neural Networks (RNNs) [1]- [3]. One of the key challenges in video captioning is extracting relevant features from videos using deep learning models like convolutional neural networks (CNNs). Some of the popular pre-trained CNN models, including VGG16, ResNet, Inception-v3, and EfficientNet, excel in image classification tasks and serve as feature extractors for video captioning. VGG16 gained early prominence with state-of-the-art results, while ResNet's residual connections tackled the vanishing gradient problem for better performance in deeper networks. Inception-v3 combines VGG16 and ResNet advantages in a deeper yet easier-to-train architecture. EfficientNet, introduced in 2019, stands

[4]https://www.tensorflow.org/api_docs/python/tf/keras/preprocessing/text/Tokenizer

out for exceptional performance and computational efficiency, employing a compound scaling method for video captioning and computer vision tasks. Researchers successfully integrate EfficientNet into captioning pipelines [13], demonstrating its effectiveness in generating accurate and descriptive video captions. Specifically, three CNN models, EfficientNB0[5],

TABLE I
MODEL PARAMETERS

| Parameter | Value |
| --- | --- |
| Frame size | (224, 224) |
| Vocabulary size | 15585 |
| time_steps_encoder | 30 (num of frames) |
| num_encoder_tokens | 1280, 2048, 4096 |
| hidden_dim | 512, 1024 |
| time_steps_decoder | 10 (max. Seq length) |
| num_decoder_tokens | 15585 (vocab_size) |

ResNet101[6], and VGG16[7], are utilized in this work, each with a different number of encoder tokens (1280, 2048, and 4096, respectively). While conducting the experiments, InceptionV3 and ResNet50 were also evaluated. However, their performance was notably lower compared to the three aforementioned models. As a result, these two models were excluded from the study. Additionally, three RNN models, namely Long Short-Term Memory (LSTM), Gated Recurrent Unit (GRU), and Bidirectional LSTM (BiLSTM), are tested, and each combination is evaluated with two different hidden dimensions (512 and 1024). The model specifications, as summarized in Table I, consist of essential parameters such as the vocabulary size, time steps for the encoder and decoder, and the number of encoder and decoder tokens.

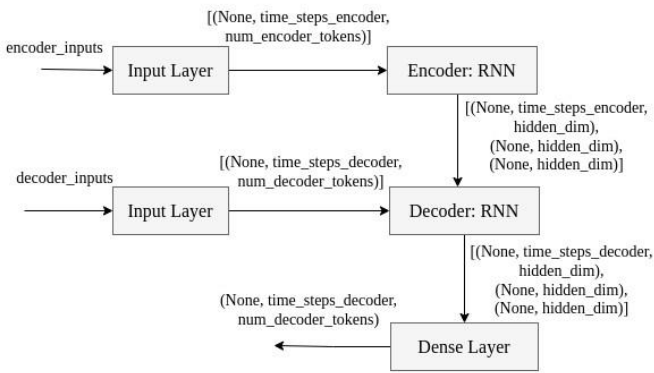

Fig. 5. Block diagram of the model architecture

During the model's implementation, the encoder takes the video features extracted by the chosen CNN model as input,

[5]https://keras.io/api/applications/efficientnet/
[6]https://keras.io/api/applications/resnet/
[7]https://keras.io/api/applications/vgg/

using the corresponding input shape explained during the video preprocessing step. The encoder processes these video features and generates hidden states, which serve as the initial states for the decoder. The decoder, in turn, processes the tokenized caption sequences and generates predicted output sequences. To predict the probability distribution over the vocabulary for each token in the caption, a Dense layer with a softmax activation function is employed as the final layer. Fig. 5 depicts the overall model architecture employed in this work.

The training process of the model involves compiling it using the 'adam' optimizer, which is a popular choice for deep learning models due to its adaptive learning rates and momentum updates. For this specific task of video captioning, the 'categorical_crossentropy' loss function is utilized. This loss function is well-suited for multi-class classification problems, like predicting words from a vocabulary in the captioning task.

Throughout the experiments, the model is trained and evaluated on the dataset with various combinations of CNN and RNN models with varying parameters. The goal is to investigate the effectiveness of different architectures in generating accurate and contextually relevant captions for Nepali videos.

D. Model Evaluation

In the model evaluation phase, the Nepali video captioning model was carefully assessed using various evaluation metrics to determine its effectiveness in generating accurate and meaningful captions in the Nepali language. It was evaluated based on metrics such as BLEU-1, BLEU-2, BLEU-3, BLEU-4 (Bilingual Evaluation Understudy), and METEOR (Metric for Evaluation of Translation with Explicit Ordering), each targeting different aspects of caption quality, such as word matching and sequence accuracy. BLEU ranges from 0 to 1, with 1 being a perfect match. METEOR also ranges from 0 to 1, with 1 representing a perfect alignment. Both metrics are widely used in evaluating machine translation quality. In our work, BLEU and METEOR scores are scaled from 0 to 100, with 100 indicating a perfect match with the reference translation. By combining these diverse metrics, a comprehensive understanding of the model's strengths and weaknesses was obtained. This analysis allowed for informed decisions regarding its suitability for real-world Nepali video captioning applications.

IV. EXPERIMENTATION AND RESULTS

The study involved experimenting with various combinations of Convolutional Neural Networks (CNNs) and Recurrent Neural Networks (RNNs) for Nepali video captioning. Different model configurations were explored by varying batch size (64, 128, 512), number of epochs (15, 30, 60), hidden dimensions (512, 1024), and the choice of CNN models, which included EfficientNetB0, ResNet101, and VGG16. The dataset comprised 1576, and 197 videos with 64477, and 8120 text descriptions for training and validation. Evaluation on the test set (197 videos with 8229 text descriptions) was performed using BLEU and METEOR scores as shown in Table II. From experiments it is found that training the model with a batch size of 128 up to 30 epochs performed better. However, models had mixed results for different hidden dimensions. Furthermore, the results revealed that the choice of CNN model significantly impacted the overall captioning accuracy. Among the tested CNNs, EfficientNetB0 consistently outperformed ResNet101 and VGG16 across most model configurations indicating its effectiveness in extracting meaningful visual features and producing contextually relevant Nepali video captions. However, it's worth noting that the choice of RNN architecture also plays a role in the model's performance. For instance, with EfficientNetB0, the combination of BiLSTM as the RNN decoder using 1024 hidden dimensions achieved the highest BLEU and METEOR scores, outperforming LSTM and GRU. On the other hand, when using ResNet101 or VGG16 as the CNN model, the choice of RNN architecture also impacted the results, with different combinations showing variations in captioning accuracy. Moreover, these findings provide valuable insights for improving video captioning performance.

TABLE II
MODEL PERFORMANCE

| Model | Hidden dim. | BLEU-1 | BLEU-2 | BLEU-3 | BLEU-4 | METEOR |
|---|---|---|---|---|---|---|
| EfficientNetB0 + LSTM | 512 | 63 | 40 | 20 | 14 | 47 |
| ResNet101 + LSTM | 512 | 56 | 33 | 16 | 12 | 44 |
| VGG16 + LSTM | 1024 | 59 | 36 | 17 | 10 | 44 |
| EfficientNetB0 + GRU | 1024 | 58 | 34 | 15 | 11 | 41 |
| ResNet101 + GRU | 512 | 52 | 29 | 9 | 6 | 39 |
| VGG16 + GRU | 1024 | 51 | 28 | 10 | 8 | 38 |
| EfficientNetB0 + BiLSTM | 1024 | 65 | 41 | 23 | 17 | 46 |
| ResNet101 + BiLSTM | 512 | 58 | 36 | 14 | 11 | 45 |
| VGG16 + BiLSTM | 1024 | 58 | 32 | 13 | 9 | 42 |

## V. DISCUSSION

The study on Nepali video captioning is an exciting area of research with several recent advancements. The experiments' results highlight the significance of model selection in achieving better performance, and EfficientNetB0 demonstrated excellent performance among the models tested. The capability of EfficientNetB0 to effectively extract crucial visual details from the videos seems to contribute to its success in generating accurate captions.. The choice of RNN also matters, with BiLSTM showing better performance compared to LSTM or GRU. BiLSTM's ability to understand both past and future information in the videos contributes to generating accurate captions.

TABLE III
MODEL COMPARISION

| Model | BLEU-4 | METEOR |
|---|---|---|
| EfficientNetB0 + LSTM (Ours) | 14 | 47 |
| EfficientNetB0 + BiLSTM (Ours) | 17 | 46 |
| VGG16 + ReBiLSTM [5] | 37.3 | 30.3 |
| ResNet152 + AC_LSTM [10] | 47.3 | 32.1 |

However, in our study, both BLEU-3 and BLEU-4 scores were comparatively low, as shown in Table III. When comparing with other works, particular attention was given to BLEU-4 and METEOR scores. One possible reason for this could be the quality of the dataset used for training and evaluation. The dataset was translated from English to Nepali using the Google Translate API, which may introduce inaccuracies and variations in the Nepali captions. These inaccuracies could affect the alignment between the generated captions and the reference sentences, leading to lower BLEU scores. Nonetheless, the study achieved good METEOR scores, indicating that the captions are meaningful and convey the main message, even if they slightly differ from the reference sentences. This highlights the model's capability to produce relevant captions despite the complexities of the Nepali language such as morphological structures, word order, multiple politeness levels, limited language resources, and potential errors introduced during translation to name a few. The study also found that the choice of batch size, hidden dimensions and number of training epochs can influence the model's performance. Further improvements can be explored by adjusting these as well as other hyperparameters. Overall, the study demonstrates that Nepali video captioning is promising and feasible. However, there is still scope for enhancing the quality of captions and improving the model.

## VI. LIMITATIONS

This study has some limitations as well. The model's capacity to generate high-quality captions for all video types may be limited. Likewise, the model's current tuning may not be optimal, leaving enough room for improvement. Additionally, the study's scope does not encompass all possible pre-trained CNN models. Moreover, the datasets used in the study are limited, and the quality of translated captions relies solely on the capabilities of Google Translate.

## VII. Conclusion and Future Works

In this work, experiments on different CNN-RNN models for Nepali video captioning were conducted. The dataset was prepared using Google Translate and preprocessed to ensure its compatibility with the study. Upon training and testing with those datasets, it was found that EfficientNetB0 + BiLSTM with 1024 hidden dimensions performed better compared to all other models used in the study. Notably, EfficientNetB0 also exhibited favorable results when used with LSTM and GRU, outperforming ResNet-101 and VGG16. Moreover, the dataset prepared in this work can be utilized for experimenting with other models for Nepali video captioning. This effort aims to provide valuable resources and serve as a reference for further research in this domain. The study can be extended by incorporating large datasets such as MSR-VTT as well as improving the quality of captions. Additionally, more experiments can be conducted on the hyperparameters to fine-tune the models effectively. Moreover, the transformer networks and other large language models like BERT are considered alternatives to traditional RNN models, which may lead to further improvements in caption generation.